\documentclass[10pt]{article}

\usepackage[
  letterpaper,
  textwidth=5.8in,
  textheight=9.0in,
  centering
]{geometry}

\usepackage[T1]{fontenc}
\usepackage[utf8]{inputenc}
\usepackage[scale=1.03]{newtxtext}
\usepackage{microtype}

\usepackage{amsmath,amssymb}
\usepackage[scale=1.03]{newtxmath}
\usepackage{graphicx}
\usepackage{booktabs}
\usepackage{multirow}
\usepackage{subcaption}
\usepackage{placeins}
\usepackage{float}
\usepackage{fvextra}
\usepackage{enumitem}
\usepackage{natbib}
\usepackage{hyperref}
\hypersetup{
  colorlinks=false,
  pdfborder={0 0 1},
  pdfborderstyle={/S/S/W 1},
  linkbordercolor={1 0 0},
  citebordercolor={0 1 0},
  urlbordercolor={0 1 1}
}
\usepackage[table]{xcolor}
\usepackage{caption}

\definecolor{lapoblue}{HTML}{2F6B9A}
\definecolor{lapogreen}{HTML}{2E7D62}
\definecolor{lapoorange}{HTML}{C56A2D}
\definecolor{lapored}{HTML}{B84A4A}
\renewenvironment{abstract}{%
  \begin{center}
    {\bfseries \abstractname\vspace{-.5em}\vspace{0pt}}
  \end{center}%
  \quote
}{%
  \endquote
}

\title{
  \textbf{LOTAPO: Leave-One-Turn Attribution for Policy Optimization\\
  with Self-Generated Process Rewards in Multi-Turn Search Reasoning}
}

\author{
  Qiang Zhu \qquad Jiajun Wu \qquad Longyi Wang\textsuperscript{*} \\
  Zhejiang University \\
  \texttt{zzhuq@zju.edu.cn} \qquad \texttt{22551345@zju.edu.cn} \qquad \texttt{longyi.24@intl.zju.edu.cn}
}

\date{}

\begin{document}

\maketitle

\begin{abstract}
Reinforcement learning for multi-turn search reasoning typically relies on terminal outcome rewards, which cannot distinguish useful, redundant, and harmful intermediate interactions.
We propose \textbf{LOTAPO}, a self-generated process-supervision method based on backward leave-one-turn attribution.
For each search turn, LOTAPO replaces the turn and its retrieval observation with a fixed \texttt{[DELETE]} placeholder and measures the resulting change in the current policy's mean log-likelihood of the gold answer.
This \emph{Answer-Likelihood Gain} estimates the turn's contribution while preserving all downstream interactions, allowing early evidence to be evaluated in the complete reasoning context.
LOTAPO further applies sign-consistency gating, retaining only normalized process advantages whose directions agree with their raw attribution scores.
The method requires no additional reward model, teacher, verifier, or LLM-as-a-Judge.
Across seven knowledge-intensive question-answering datasets with local retrieval, LOTAPO achieves an average exact-match score of \(0.326\), outperforming the strongest step-reward baseline, IGPO, by \(0.053\).
Ablations show complementary benefits from backward attribution and sign-consistency gating, demonstrating that policy-derived retrospective attribution can provide effective process supervision for multi-turn search agents.
Code is available at \url{https://github.com/zhuq-111/LOTAPO-Leave-One-Turn-Attribution}.
\end{abstract}

\section{Introduction}

Large language models increasingly need to interact with external information sources to solve knowledge-intensive problems rather than relying solely on parametric knowledge.
In multi-turn search reasoning, a model must iteratively formulate queries, retrieve information, interpret observations, and integrate evidence before producing an answer.
Reinforcement learning (RL) provides a natural training framework for this setting: it can optimize complete interaction trajectories according to task-level objectives without requiring human demonstrations for every intermediate action~\cite{jin2025searchr1}.
However, the long-horizon nature of search reasoning creates a key credit-assignment problem: the final answer usually depends on multiple interrelated search and reasoning decisions, while both successful and failed trajectories may contain useful, redundant, and harmful intermediate interactions.

When training relies only on final-answer correctness, all actions within a trajectory are typically updated according to the same outcome.
Consequently, a successful trajectory may incorrectly reinforce ineffective searches, while a failed trajectory may penalize a turn that retrieved crucial evidence but whose evidence was not used correctly in subsequent reasoning.
A terminal reward indicates whether the entire trajectory succeeded, but it does not reveal the specific attribution of each intermediate interaction to the final outcome.
Multi-turn search reasoning therefore calls for finer-grained process-level credit signals.

Existing methods commonly obtain process supervision from reward models, teacher models, verifiers, or other step-level evaluators~\cite{lightman2023verify,wang2024mathshepherd,fang2026ropd}~\cite{uesato2022processfeedback,zhang2024restmcts}.
Although these methods provide denser feedback, they introduce additional training and inference costs and make policy optimization depend on the quality of an external evaluator.
Moreover, an evaluator trained on another policy or data distribution may not accurately assess the intermediate behaviors produced by the current policy.
This raises a natural question: can process-level supervision be constructed directly from signals produced by the current policy?

To this end, we propose \textbf{LOTAPO}, a method for self-generated process rewards based on backward leave-one-turn attribution.
Given a complete trajectory, LOTAPO replaces each eligible search turn and its retrieval observation in turn with \texttt{[DELETE]}, and measures the resulting change in the current policy's mean log-likelihood of the gold answer.
We call this change the \emph{Answer-Likelihood Gain} and use it to estimate the attribution of each search turn within the complete reasoning process.
Unlike local confidence changes between adjacent states, LOTAPO preserves all remaining downstream interactions when evaluating a target turn, thereby identifying early evidence whose value becomes apparent only through subsequent reasoning.

Raw answer-likelihood gains, however, can be noisy and can vary in scale across sampled trajectories.
LOTAPO applies robust scaling and group normalization, then introduces sign-consistency gating to retain only process signals whose normalized advantages agree in direction with their raw answer-likelihood gains.
The gated process advantage is assigned to the policy-generated tokens in the corresponding search turn and combined with the terminal outcome advantage used by GRPO~\cite{shao2024deepseekmath}.
The final-answer turn remains supervised only by the terminal reward.
The entire procedure derives supervision directly from the current policy and requires no separately trained process reward model, teacher model, verifier, or LLM-based judge.

Our main contributions are as follows:
\begin{itemize}
    \item We identify a limitation of outcome rewards for within-trajectory credit assignment in multi-turn search reasoning: final-answer correctness does not reflect the specific attribution of individual intermediate search interactions.

    \item We propose LOTAPO, a self-generated process-reward method based on backward leave-one-turn attribution. LOTAPO estimates turn-level attribution from changes in the current policy's likelihood of the gold answer without relying on an additional process evaluator.

    \item We introduce group-relative sign-consistency gating to filter directionally unreliable attribution signals, and evaluate LOTAPO on seven knowledge-intensive question-answering datasets.
\end{itemize}

\section{Related Work}
\label{sec:related_work}

\subsection{Reinforcement Learning for Search-Augmented Reasoning}

Recent work uses reinforcement learning to train language models to actively acquire external information during reasoning.
Unlike conventional methods that treat retrieval as a fixed preprocessing step, these approaches allow a model to decide when to search, how to formulate a query, and how to integrate the retrieved evidence.
Search-R1 shows that outcome-reward-based reinforcement learning can induce multi-turn search behavior without supervised reasoning trajectories~\cite{jin2025searchr1}.
R1-Searcher and ReSearch likewise use task-level outcome rewards to train search and reasoning capabilities~\cite{song2025r1searcher,chen2025research}.
This paradigm has also been extended to web navigation, multimodal search, and retrieval-augmented reasoning, as in WebAgent-R1, MMSearch-R1, and RAG-R1~\cite{wei2025webagentr1,wu2026mmsearchr1,tan2026ragr1}.
In addition to answer correctness, these methods may incorporate auxiliary signals such as format rewards, tool-use constraints, search costs, or efficiency objectives.

These studies indicate that trajectory-level reinforcement learning can induce complex search behavior.
However, trajectory-level rewards primarily reflect whether the complete interaction succeeds and provide limited information about which intermediate queries, retrieval observations, or reasoning turns contribute to the final outcome.
LOTAPO retains terminal task supervision while extracting turn-level attribution signals from complete trajectories, providing finer-grained process supervision for this setting.

\subsection{External Process Supervision and On-Policy Distillation}

One line of work introduces denser supervision through external process evaluators.
Let's Verify Step by Step trains a process reward model using human step-level annotations~\cite{lightman2023verify}.
Math-Shepherd reduces dependence on human annotation through continuation-based automatic verification~\cite{wang2024mathshepherd}.
ReST-MCTS* combines process-reward estimation with tree search and trains on the probability that an intermediate state eventually leads to a correct answer; CriticSearch instead uses a frozen retrospective critic to generate turn-level rewards from the complete trajectory and gold answer~\cite{zhang2024restmcts,zhang2026criticsearch}.
These methods evaluate intermediate reasoning using human annotation, automatic verification, search-derived value estimates, or external critics.

Recent work further uses strong language models or structured semantic information to supervise trajectories generated by the current policy.
Rubric-based On-policy Distillation (ROPD) induces prompt-specific rubrics from teacher--student discrepancies and uses them to evaluate on-policy student responses~\cite{fang2026ropd}.
OPID extracts turn-level and step-level skills from complete trajectories and constructs skill-conditioned distillation signals for agent reinforcement learning~\cite{yang2026opid}.

Externally guided process supervision can provide semantically rich feedback and distinguish among different reasoning or planning errors.
However, such methods generally require additional teacher calls, reward models, rubric induction, skill extraction, or verification procedures, and their reliability depends on the capability and calibration of the external evaluator.
In contrast, LOTAPO neither generates semantic critiques or rubrics nor extracts reusable skills.
Instead, it estimates the conditional attribution of each search turn using the current policy's likelihood of the gold answer.
LOTAPO therefore requires no separately learned process evaluator; its process signal depends only on terminal task supervision and the current policy's own probabilistic estimates.

\subsection{Policy-Derived Process Rewards and Turn-Level Credit Assignment}

The work most closely related to LOTAPO constructs intermediate rewards from changes in the policy's own predictions or interaction states.
IGPO and TIPS both generate turn-level rewards from increments in gold-answer likelihood between adjacent states; TIPS uses a frozen, periodically updated copy of the policy to compute an information potential function~\cite{wang2026igpo,xie2026tips}.
StepSearch combines information gain, redundancy control, and token-level process supervision, while IG-Search estimates the contribution of retrieval evidence by comparing answer confidence under the retrieved documents and randomly substituted documents~\cite{zheng2025stepsearch,liang2026igsearch}.
Other work measures reasoning progress through semantic information gain, reward propagation, or turn-level value estimation~\cite{hu2026inforeasoner,feng2026rewardflow,wei2025turnlevel}.

Overall, among policy-derived approaches that do not rely on external critics, prior work mainly estimates reasoning progress through changes between adjacent states or local counterfactual comparisons.
LOTAPO instead emphasizes retrospective leave-one-turn attribution over the complete trajectory.
Forward methods typically compare adjacent states before and after adding the current interaction:

\begin{equation}
s(y^\star \mid c_{\leq t})
-
s(y^\star \mid c_{<t}),
\end{equation}
which measures the immediate effect of the current interaction on the model's likelihood of generating the gold answer.
In multi-hop search, however, evidence obtained early may become useful only when combined with later observations, so a local state change may not accurately capture its attribution.

LOTAPO therefore performs backward leave-one-turn attribution over the complete trajectory:
\begin{equation}
s(y^\star \mid c_{\mathrm{full}})
-
s(y^\star \mid c_{\setminus t}),
\end{equation}
where the target search turn and its retrieval observation are jointly replaced with \texttt{[DELETE]}, while all other downstream interactions remain unchanged.
This design evaluates the attribution of an early search turn in the full reasoning context.
Unlike IG-Search, which performs document-level attribution by substituting retrieved documents, LOTAPO uses an entire search interaction as its attribution unit and thereby provides turn-level process supervision for multi-turn search reasoning.

\section{Method}
\label{sec:method}

\begin{figure}[H]
\centering
\includegraphics[width=\linewidth]{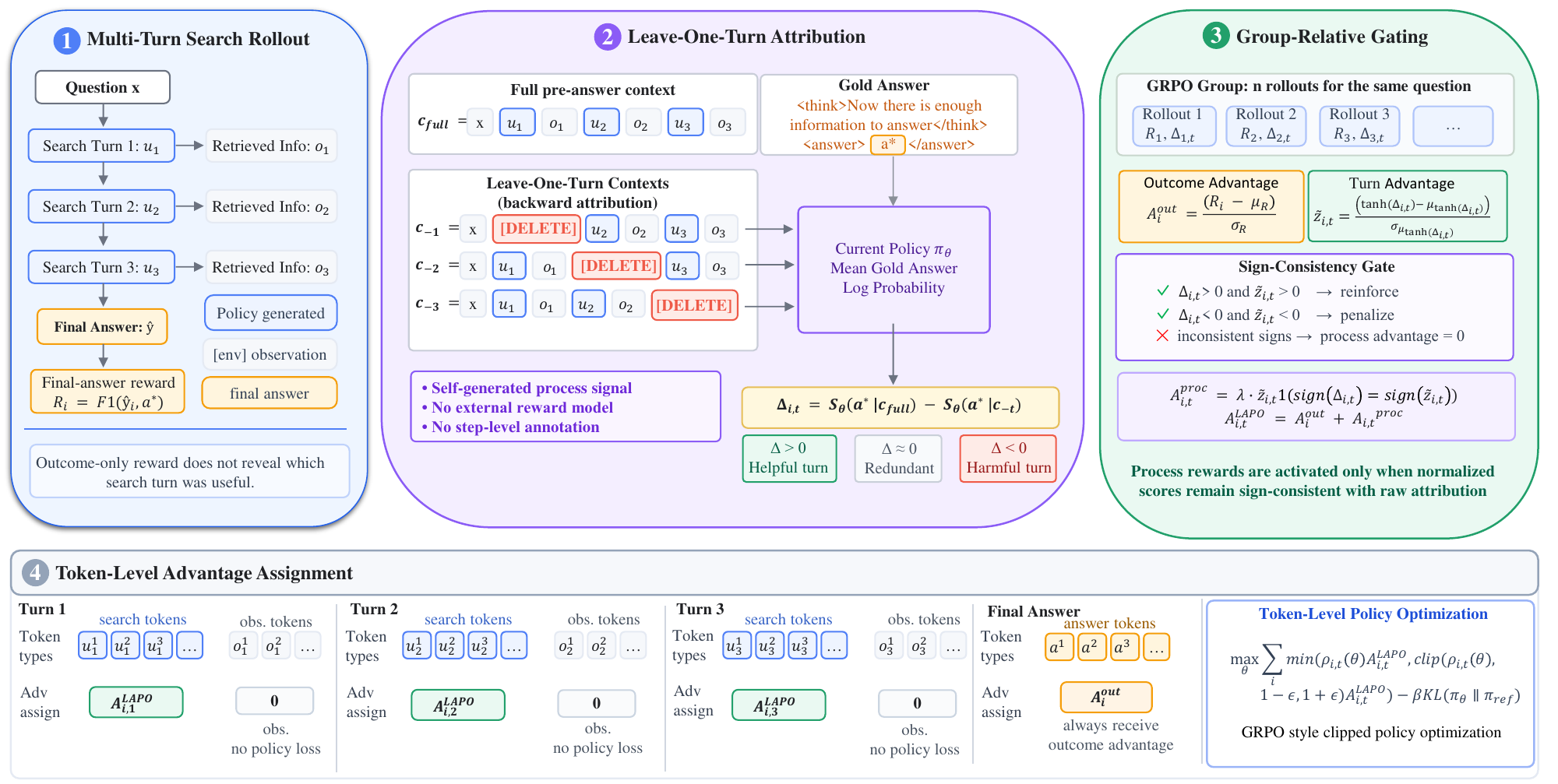}
\caption{\textbf{Overview of LOTAPO.}
For each complete search trajectory, LOTAPO replaces each eligible search turn and its retrieval observation in turn with \texttt{[DELETE]}, and estimates turn-level attribution using the difference in the current policy's mean log-likelihood of the gold answer before and after replacement.
The resulting attribution scores undergo robust scaling, a tanh bounded transformation, group normalization, and sign-consistency gating.
They are then assigned to the policy-generated tokens of the corresponding search turns and used jointly with the trajectory-level outcome advantage for policy optimization.}
\label{fig:lapo-overview}
\end{figure}

\subsection{Problem Formulation and Method Overview}

We consider a multi-turn search agent trained with reinforcement learning.
Given a question $x_i$, the policy $\pi_\theta$ alternates between generating search actions and receiving observations from the retrieval environment, and eventually generates an answer.
A complete trajectory is represented as
\begin{equation}
\tau_i
=
\left(
x_i,
u_{i,1},o_{i,1},
\ldots,
u_{i,T_i},o_{i,T_i},
v_i
\right),
\end{equation}
where $u_{i,t}$ denotes the policy-generated \texttt{<think>...<search>...</search>} content at search turn $t$, $o_{i,t}$ denotes the corresponding \texttt{<information>...</information>} retrieval observation returned by the environment, $T_i$ is the number of search turns, and $v_i$ denotes the policy-generated final-answer turn \texttt{<think>...<answer>...</answer>}.
We use $\hat{a}_i$ to denote the final answer extracted from $v_i$.
We treat a search action and its corresponding retrieval observation as one complete search interaction:
\begin{equation}
e_{i,t}=(u_{i,t},o_{i,t}).
\end{equation}
A \emph{valid search turn} is a search interaction that contains both a complete policy-generated search action and its corresponding environment observation; the final-answer turn is not a valid search turn.

Let $a_i^\star$ denote the gold answer to question $x_i$.
A conventional outcome reward provides a uniform training signal for the entire trajectory based on final-answer quality.
It cannot distinguish the specific attribution of different search turns and thus offers limited fine-grained credit assignment for crucial, redundant, or misleading search behavior.

LOTAPO addresses this limitation through backward leave-one-turn attribution.
It measures the change in the current policy's conditional likelihood of the gold answer when a single search turn is replaced in its entirety with \texttt{[DELETE]}, and uses the change to construct a turn-level process advantage.
After robust scaling, a tanh bounded transformation, group normalization, and sign-consistency gating, this process advantage is assigned to the policy-generated tokens of the corresponding search turn and combined with the original trajectory-level outcome advantage in GRPO optimization.

\subsection{Backward Leave-One-Turn Attribution}

\paragraph{Gold-answer likelihood.}

Let $c$ denote the context before the policy generates its final answer.
To measure the current policy's support for the gold answer, we append a fixed response template containing the gold answer to $c$ and score only the gold-answer tokens in the template.

Let $p$ denote the fixed token sequence in the scoring template that precedes the gold-answer tokens.
Write
$a_i^\star=(a_{i,1}^\star,\ldots,a_{i,L_i}^\star)$,
where $L_i$ is the number of tokens in the gold answer.
The current policy's score for the gold answer is
\begin{equation}
S_\theta(a_i^\star\mid c)
=
\frac{1}{L_i}
\sum_{k=1}^{L_i}
\log
\pi_\theta
\left(
a_{i,k}^\star
\mid
c,p,a_{i,<k}^\star
\right).
\label{eq:gold_score}
\end{equation}
The mean token log-likelihood reduces the effect of answer length on the score scale, making attribution scores more comparable across questions.

Let $c_i^{\mathrm{full}}$ denote the pre-answer context that contains all search interactions but excludes the policy-generated final-answer turn $v_i$.
Excluding $v_i$ prevents the policy's original final reasoning and answer from leaking into gold-answer scoring, so the score reflects only the support supplied by the preceding search interactions.

\paragraph{Leave-one-turn counterfactual attribution.}

For each valid search turn $t$, LOTAPO applies a leave-one-turn operation to the corresponding search action and retrieval observation
$e_{i,t}=(u_{i,t},o_{i,t})$ in the full context $c_i^{\mathrm{full}}$.
Specifically, we replace the entire target turn with a fixed \texttt{[DELETE]} placeholder to explicitly mark the absence of that turn's content in the counterfactual context.
This produces the counterfactual context $c_{i,-t}$, in which every other search interaction and its order remain unchanged.

The answer-likelihood gain of search turn $t$ is defined as
\begin{equation}
\Delta_{i,t}
=
S_\theta(a_i^\star\mid c_i^{\mathrm{full}})
-
S_\theta(a_i^\star\mid c_{i,-t}).
\label{eq:raw_attribution}
\end{equation}

When $\Delta_{i,t}>0$, replacing the target turn with the placeholder reduces the gold-answer likelihood, indicating positive attribution to that turn under the current context and policy.
When $\Delta_{i,t}<0$, the replacement instead increases the gold-answer likelihood, suggesting that the turn may have a negative influence under the current context and policy.
A value of $\Delta_{i,t}$ near zero indicates that the turn has little effect on the gold-answer likelihood.

Unlike forward attribution methods that measure an immediate change by comparing adjacent prefix states, LOTAPO preserves all downstream search interactions when evaluating the target turn.
This backward leave-one-turn design enables retrospective attribution in the full reasoning context even when evidence from an early turn becomes useful only through later search and reasoning.

In our implementation, we flatten the full contexts and all leave-one-turn counterfactual contexts in a rollout batch into a single scoring batch, and append the same gold-answer scoring template to every context.
A batched forward pass produces the gold-answer log-likelihoods under the full and counterfactual contexts simultaneously, reducing separate model calls and scheduling overhead.

\subsection{Gated Process-Advantage Construction and Assignment}

\paragraph{Trajectory-level outcome advantage.}

For each question, we sample $G$ trajectories from the current policy.
The terminal task reward for trajectory $i$ is
\begin{equation}
R_i
=
\operatorname{F1}(\hat{a}_i,a_i^\star).
\end{equation}
For a trajectory group $g$ corresponding to the same question, the outcome advantage is obtained by within-group standardization:
\begin{equation}
A_i^{\mathrm{out}}
=
\frac{
R_i-\mu_R^{(g)}
}{
\sigma_R^{(g)}+\epsilon_R
},
\label{eq:outcome_advantage}
\end{equation}
where $\mu_R^{(g)}$ and $\sigma_R^{(g)}$ are the mean and standard deviation of rewards within the group, respectively, and $\epsilon_R$ is included for numerical stability.
This outcome advantage provides terminal task supervision to every policy-generated token in the trajectory and remains the base advantage after LOTAPO introduces turn-level credit signals.

\paragraph{Robust scaling and group normalization.}

Answer-likelihood gains may differ in scale across questions and trajectories, and a small number of extreme values may destabilize policy updates.
Let $\mathcal{E}_g$ be the set of all valid trajectory--turn pairs in group $g$.
LOTAPO estimates a scale factor using the median nonzero absolute answer-likelihood gain within the group:
\begin{equation}
\mathcal{D}_g
=
\left\{
|\Delta_{j,s}|:
(j,s)\in\mathcal{E}_g,\,
|\Delta_{j,s}|>\epsilon_{\mathrm{ig}}
\right\},
\qquad
\tau_g
=
\begin{cases}
\operatorname{median}(\mathcal{D}_g)+\epsilon_{\mathrm{ig}},
& \mathcal{D}_g\neq\varnothing,\\
\epsilon_{\mathrm{ig}},
& \mathcal{D}_g=\varnothing.
\end{cases}
\label{eq:robust_scale}
\end{equation}
Here, $\epsilon_{\mathrm{ig}}$ is a numerical-stability constant.
When all answer-likelihood gains in a group are zero, $\mathcal{D}_g$ is empty and $\tau_g$ falls back to $\epsilon_{\mathrm{ig}}$.
All bounded gains $z_{i,t}$ are then zero, so LOTAPO adds no process advantage.

We next apply a bounded transformation to the answer-likelihood gain using the hyperbolic tangent:
\begin{equation}
z_{i,t}
=
\tanh
\left(
\frac{\Delta_{i,t}}{\tau_g}
\right).
\label{eq:bounded_attribution}
\end{equation}
This transformation preserves the sign and relative magnitude of the raw answer-likelihood gain while limiting the influence of extreme values on subsequent advantage computation.

Finally, LOTAPO standardizes the bounded gains within the group:
\begin{equation}
\widetilde{z}_{i,t}
=
\frac{
z_{i,t}-\mu_z^{(g)}
}{
\sigma_z^{(g)}+\epsilon_z
},
\label{eq:normalized_attribution}
\end{equation}
where $\mu_z^{(g)}$ and $\sigma_z^{(g)}$ are the mean and standard deviation, respectively, of
$\{z_{j,s}:(j,s)\in\mathcal{E}_g\}$.
This normalization spans all sampled trajectories for the same question and all valid search turns in those trajectories.
Thus, $\widetilde{z}_{i,t}$ measures the target turn's contribution relative to the other valid turns in the same group and provides a scale-aligned process signal across questions and trajectories.

\paragraph{Sign-consistency gating and advantage assignment.}

Although group normalization measures the relative contribution of a search turn, subtracting the group mean can reverse the sign of the raw answer-likelihood gain.
For example, a turn with a negative raw attribution may receive a positive normalized advantage if its contribution is above the group mean.
Using such a directionally reversed signal directly could make the policy update contradict the raw attribution identified by the leave-one-turn comparison.

LOTAPO therefore introduces sign-consistency gating, which retains only process signals for which the raw answer-likelihood gain and normalized turn advantage point in the same direction:
\begin{equation}
g_{i,t}
=
\mathbb{I}
\left[
\Delta_{i,t}\widetilde{z}_{i,t}>0
\right].
\label{eq:consistency_gate}
\end{equation}
The gated process advantage is
\begin{equation}
A_{i,t}^{\mathrm{proc}}
=
g_{i,t}\widetilde{z}_{i,t}.
\label{eq:process_advantage}
\end{equation}

When $\Delta_{i,t}$ and $\widetilde{z}_{i,t}$ have the same sign, the normalized turn advantage is retained; when their directions disagree, the process advantage is set to zero.
This gate filters only directionally unreliable auxiliary process signals and does not affect the trajectory's original outcome advantage.

LOTAPO then assigns the process advantage to the policy-generated tokens in the corresponding search turn.
For the $k$-th policy-generated token in trajectory $i$, the training advantage is
\begin{equation}
A_{i,k}^{\mathrm{token}}
=
\begin{cases}
A_i^{\mathrm{out}}
+
\lambda A_{i,t}^{\mathrm{proc}},
&
k \text{ belongs to valid search turn }t,
\\
A_i^{\mathrm{out}},
&
k \text{ belongs to the final-answer turn},
\end{cases}
\label{eq:token_advantage}
\end{equation}
where $\lambda$ controls the weight of the process advantage.
Thus, all policy-generated tokens retain terminal task supervision, while search turns that pass sign-consistency gating receive an additional turn-level process signal.
The final-answer turn participates in neither forward nor backward process-reward computation and receives only the trajectory-level outcome advantage.
Retrieval observations are generated by the environment and do not participate in policy optimization.

\paragraph{Policy optimization.}

LOTAPO directly uses the clipped surrogate objective of GRPO for policy updates.
Let $\pi_{\theta_{\mathrm{old}}}$ be the old policy that generated the current trajectories, and let $h_{i,k}$ be the complete context before the $k$-th policy token, including the question, all preceding policy tokens, and environment observations.
The importance-sampling ratio for the $k$-th policy token is
\begin{equation}
\rho_{i,k}(\theta)
=
\frac{
\pi_\theta(y_{i,k}\mid h_{i,k})
}{
\pi_{\theta_{\mathrm{old}}}(y_{i,k}\mid h_{i,k})
}.
\label{eq:importance_ratio}
\end{equation}

Using the token-level advantage above, the LOTAPO objective is
\begin{align}
\mathcal{J}_{\mathrm{LOTAPO}}(\theta)
&=
\mathbb{E}_{x\sim\mathcal{D},\,
\tau_i\sim\pi_{\theta_{\mathrm{old}}}}
\Bigg[
\frac{1}{|\mathcal{P}_i|}
\sum_{k\in\mathcal{P}_i}
\Bigg(
\nonumber\\[-0.2em]
&\quad
\min\Big[
\rho_{i,k}(\theta)A_{i,k}^{\mathrm{token}},
\nonumber\\[-0.2em]
&\qquad
\operatorname{clip}
\big(
\rho_{i,k}(\theta),
1-\epsilon_{\mathrm{clip}},
1+\epsilon_{\mathrm{clip}}
\big)
A_{i,k}^{\mathrm{token}}
\Big]
-
\beta
D_{\mathrm{KL}}^{(i,k)}
\big(
\pi_\theta
\Vert
\pi_{\mathrm{ref}}
\big)
\Bigg)
\Bigg],
\label{eq:lapo_objective}
\end{align}
where $\mathcal{P}_i$ is the set of policy-generated tokens in trajectory $i$ that participate in policy optimization, $\epsilon_{\mathrm{clip}}$ is the clipping range, $\pi_{\mathrm{ref}}$ is the reference policy, $D_{\mathrm{KL}}^{(i,k)}$ is the KL-regularization estimate at the $k$-th policy-token position, and $\beta$ controls the strength of KL regularization.
Training maximizes $\mathcal{J}_{\mathrm{LOTAPO}}(\theta)$.

LOTAPO does not change the form of GRPO policy optimization.
Before optimization, it uses backward leave-one-turn attribution to refine the trajectory-level outcome advantage into search-turn-specific token-level advantages.
Turn-level credit signals can therefore be integrated directly into the existing GRPO training pipeline without an additional process reward model or changes to the underlying optimizer.

\section{Experiments}
\label{sec:experiments}

\subsection{Experimental Setup}
\label{sec:experimental_setup}

\paragraph{Datasets and evaluation.}
We construct the reinforcement-learning training set from Natural Questions (NQ), TriviaQA (TQ), HotpotQA, and 2WikiMultiHopQA (2Wiki)~\cite{kwiatkowski2019naturalquestions,joshi2017triviaqa,yang2018hotpotqa,ho2020twowiki}.
Specifically, we randomly sample 20,000 examples from the training split of each dataset, yielding 80,000 training examples in total.
For each of the seven evaluation datasets below, we randomly sample 500 examples from the corresponding original split for validation; for testing, we use each dataset's test or validation split.
We evaluate all methods on seven question-answering benchmarks: NQ, TriviaQA, HotpotQA, 2Wiki, MuSiQue, Bamboogle, and PopQA~\cite{trivedi2022musique,press2023compositionality,mallen2023popqa}.
Because NQ, TriviaQA, HotpotQA, and 2Wiki are used during training, we treat them as in-domain datasets.
MuSiQue, Bamboogle, and PopQA are not used for training and evaluate out-of-domain generalization.
Together, these datasets cover both single-hop knowledge-intensive question answering and multi-hop question answering that requires iterative retrieval and evidence aggregation.
At test time, we use exact match (EM) as the primary evaluation metric.
\textit{Avg.} in the tables denotes the macro-average EM across the seven datasets, with equal weight assigned to each dataset.
During training, we use answer-level F1 as the outcome-reward signal, measuring token overlap between the model prediction and the reference answer.

\paragraph{Baselines.}
We compare LOTAPO with three categories of representative methods.
The first category consists of \emph{prompt-based methods} without task-level reinforcement-learning optimization: Direct Inference, Chain-of-Thought (CoT)~\cite{wei2022chainofthought}, Search-o1~\cite{li2025searcho1}, and retrieval-augmented generation (RAG)~\cite{lewis2020rag}.
These methods represent direct answering, explicit reasoning, and retrieval-augmented paradigms.
The second category comprises \emph{outcome-reward-based methods}: Search-R1-base, Search-R1-instruct~\cite{jin2025searchr1}, and R1-Searcher~\cite{song2025r1searcher}.
These methods optimize entire search trajectories primarily according to final-answer quality, without explicitly modeling the separate attribution of intermediate search turns to the final outcome.
The third category comprises \emph{step-reward-based methods}: StepSearch-base, StepSearch-instruct~\cite{zheng2025stepsearch}, and IGPO~\cite{wang2026igpo}.
IGPO is the method most closely related to LOTAPO: both optimize multi-turn search behavior using process rewards, but they use different attribution mechanisms.
In the main results, IGPO denotes the original forward-attribution method without our sign-consistency gating; the ablation study separately reports a variant that adds this gate to IGPO.

\paragraph{Experimental configuration.}
All methods use a local retrieval corpus constructed from the 2018 English Wikipedia.
The corpus is processed following Karpukhin et al.~\cite{karpukhin2020dense}, and E5~\cite{wang2022e5} is used as the dense retriever.
Each search returns the three most relevant passages.
The model may perform at most three search turns before producing its final answer.
All methods share the same retrieval corpus, retriever, number of returned documents, and maximum number of search turns to prevent differences in retrieval environments from confounding the comparison.
All reinforcement-learning methods are trained for 200 optimization steps on eight NVIDIA A100 GPUs.
The global training batch size is 512, and \(G=5\) rollouts are sampled for each input prompt.
Unless otherwise specified, all methods use the same model initialization, training data, prompt template, decoding parameters, and retrieval budget.
The optimizer, learning rate, data preprocessing, and other generation settings are provided in Appendix~\ref{app:implementation_details}.

\subsection{Main Results}
\label{sec:main_results}

\begin{table}[t]
\centering
\small
\setlength{\tabcolsep}{4.2pt}
\caption{
\textbf{Exact match (EM) on seven knowledge-intensive question-answering datasets.}
NQ, TriviaQA, HotpotQA, and 2Wiki are in-domain datasets, while MuSiQue, Bamboogle, and PopQA are out-of-domain datasets.
The average is the macro-average across all seven datasets.
The best result in each column is shown in bold.
}
\label{tab:main-results}
\begin{tabular}{lcccccccc}
\toprule
\multirow{2}{*}{Method}
& \multicolumn{4}{c}{In-Domain}
& \multicolumn{3}{c}{Out-of-Domain}
& \multirow{2}{*}{Avg.} \\
\cmidrule(lr){2-5}
\cmidrule(lr){6-8}
& NQ & TQ & HotpotQA & 2Wiki
& MuSiQue & Bamboogle & PopQA & \\
\midrule
\rowcolor{gray!12}
\multicolumn{9}{l}{\textbf{Prompt-Based Methods}} \\
Direct Inference
& 0.100 & 0.224 & 0.138 & 0.200 & 0.034 & 0.036 & 0.064 & 0.114 \\
CoT
& 0.022 & 0.024 & 0.020 & 0.018 & 0.018 & 0.102 & 0.046 & 0.036 \\
Search-o1
& 0.224 & 0.368 & 0.204 & 0.180 & 0.090 & 0.282 & 0.154 & 0.215 \\
RAG
& 0.328 & 0.424 & 0.234 & 0.186 & 0.078 & 0.120 & 0.178 & 0.221 \\
\midrule
\rowcolor{gray!12}
\multicolumn{9}{l}{\textbf{Outcome-Reward-Based Methods}} \\
Search-R1-base
& 0.342 & 0.424 & 0.288 & 0.246 & 0.110 & 0.196 & 0.202 & 0.258 \\
Search-R1-instruct
& 0.356 & 0.454 & 0.268 & 0.238 & 0.130 & 0.208 & 0.184 & 0.263 \\
R1-Searcher
& 0.274 & 0.428 & 0.234 & 0.202 & 0.094 & 0.212 & 0.114 & 0.223 \\
\midrule
\rowcolor{gray!12}
\multicolumn{9}{l}{\textbf{Step-Reward-Based Methods}} \\
StepSearch-base
& 0.312 & 0.460 & 0.258 & 0.152 & 0.134 & 0.188 & 0.140 & 0.235 \\
StepSearch-instruct
& 0.286 & 0.424 & 0.262 & 0.146 & 0.130 & 0.174 & 0.120 & 0.220 \\
IGPO
& 0.358 & 0.472 & 0.308 & 0.244 & 0.136 & 0.252 & 0.144 & 0.273 \\
\textbf{LOTAPO}
& \textbf{0.422} & \textbf{0.506} & \textbf{0.338} & \textbf{0.320}
& \textbf{0.188} & \textbf{0.304} & \textbf{0.206} & \textbf{0.326} \\
\bottomrule
\end{tabular}
\end{table}

\paragraph{Overall performance.}
As shown in Table~\ref{tab:main-results}, LOTAPO achieves the best result on all seven evaluation datasets, with an average EM of \(0.326\).
LOTAPO yields an absolute improvement of \(0.063\) over the strongest outcome-reward baseline, Search-R1-instruct, and improves average EM by \(0.053\) over the strongest step-reward baseline, IGPO, corresponding to a relative gain of \(19.4\%\).

This result supports the view that final-answer feedback alone may be insufficient for credit assignment in multi-turn search trajectories.
Outcome rewards can encourage the policy to generate trajectories that succeed overall, but they do not further identify which intermediate interactions actually support the final answer.
LOTAPO converts the prediction change caused by replacing a single search turn with \texttt{[DELETE]} into a turn-level learning signal, providing finer-grained supervision without an external process reward model.

\paragraph{Comparison with step-level attribution methods.}
IGPO generally outperforms most prompt-based and outcome-reward-based methods, indicating the value of process-level supervision for training multi-turn search agents.
Building on this observation, LOTAPO achieves higher EM than IGPO on all seven datasets.

The main difference between the two methods lies in the context used to evaluate the attribution of an intermediate turn.
Forward information gain is computed immediately after the current turn is generated and therefore cannot account for evidence retrieved later.
LOTAPO instead reevaluates each intermediate turn after the complete search trajectory has been generated, using the final downstream context.
In multi-turn retrieval, the value of an early query or an intermediate piece of evidence may become apparent only after it is combined with later evidence.
Retrospective attribution can therefore capture cross-turn dependencies that are difficult to identify from a local prefix alone.
LOTAPO's clear gains on multi-hop datasets such as HotpotQA and 2Wiki are consistent with this motivation.

\paragraph{In-domain and out-of-domain generalization.}
LOTAPO's advantage also transfers to data distributions not observed during training.
Across MuSiQue, Bamboogle, and PopQA, LOTAPO attains an average EM of \(0.233\), compared with \(0.177\) for IGPO.
The improvement appears on all three out-of-domain datasets and is not driven by an anomalous gain on a single dataset.
These results indicate that LOTAPO's performance advantage transfers to the three out-of-domain datasets considered here, suggesting some cross-dataset generalization in the learned search and evidence-use behavior.

\subsection{Training Dynamics and Process-Reward Analysis}
\label{sec:training_analysis}

\begin{figure}[t]
    \centering
    \captionsetup{font=small,skip=3pt}
    \includegraphics[width=0.92\linewidth]{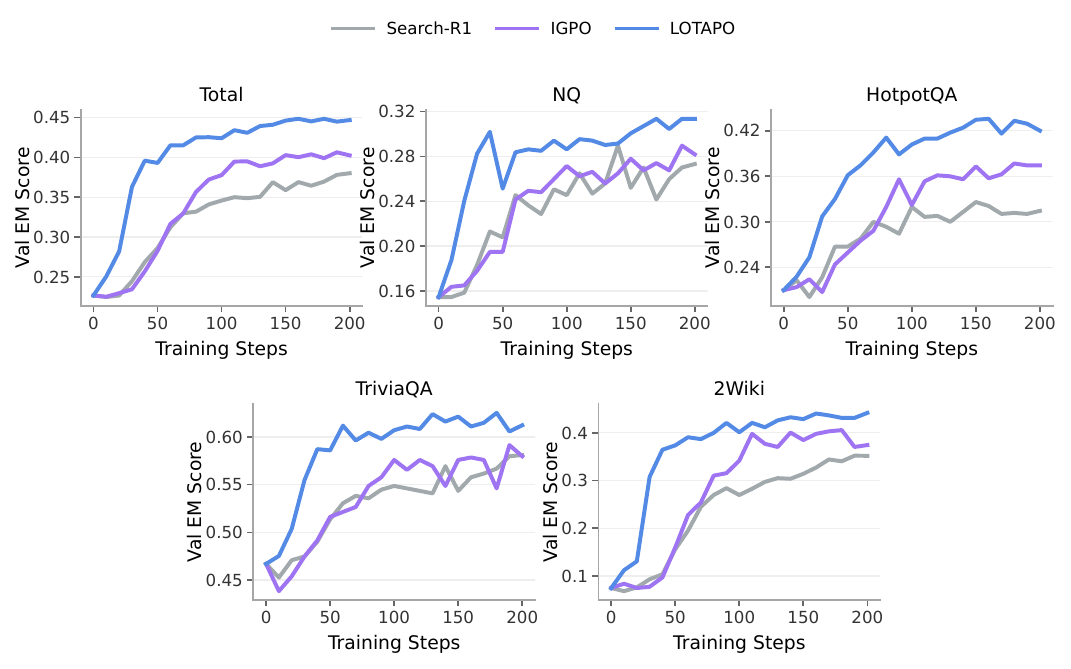}
    \caption{\textbf{Performance dynamics on held-out validation sets.}
    Exact match (EM) for LOTAPO, IGPO, and Search-R1 over training steps; Total denotes the aggregate result over the four in-domain validation sets.}
    \label{fig:validation-curves}

    \vspace{6pt}
    \begin{minipage}[t]{0.48\linewidth}
        \vspace{0pt}
        \centering
        \captionsetup{font=footnotesize,skip=2pt,justification=raggedright,singlelinecheck=false}
        \makebox[\linewidth][c]{\includegraphics[height=0.58\linewidth]{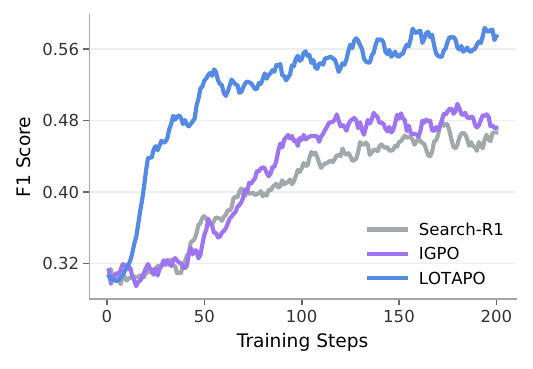}}
        \caption{\textbf{Training-performance dynamics.} Trajectory-level answer F1 for the three methods over training steps.}
        \label{fig:train-outcome}
    \end{minipage}\hfill
    \begin{minipage}[t]{0.48\linewidth}
        \vspace{0pt}
        \centering
        \captionsetup{font=footnotesize,skip=2pt,justification=raggedright,singlelinecheck=false}
        \makebox[\linewidth][c]{\includegraphics[height=0.58\linewidth]{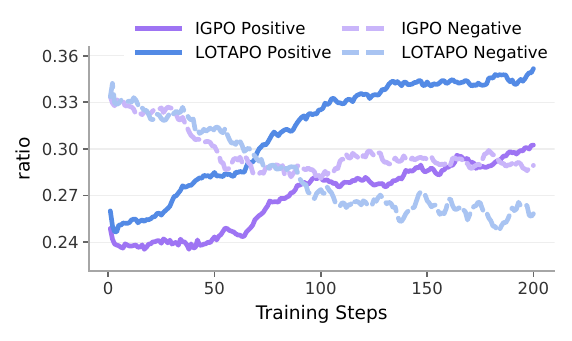}}
        \caption{\textbf{Evolution of process signals.} Proportions of turns with positive and negative answer-likelihood gains under LOTAPO and IGPO.}
        \label{fig:process-ratio}
    \end{minipage}
\end{figure}

\paragraph{Optimization dynamics and validation performance.}
Figures~\ref{fig:train-outcome} and~\ref{fig:validation-curves} compare the reinforcement-learning dynamics and held-out performance of LOTAPO, IGPO, and Search-R1.
LOTAPO improves rapidly early in training and subsequently maintains a higher trajectory-level answer F1.
Late in training, LOTAPO's training F1 approaches \(0.57\), while IGPO and Search-R1 remain around \(0.47\)--\(0.48\).
This pattern is consistent with LOTAPO's motivation of providing finer-grained turn-level supervision.

LOTAPO's advantage on training trajectories also transfers to held-out validation examples.
On the aggregate validation set, LOTAPO reaches the highest validation performance late in training and shows particularly clear advantages on HotpotQA and 2Wiki.
The training and validation curves exhibit consistent performance trends, indicating that higher outcome rewards on training trajectories accompany better validation performance on held-out queries; these curves alone, however, cannot rule out other factors such as overfitting.

In contrast, the differences among methods on TriviaQA are smaller and the curves overlap more strongly.
A possible explanation is that many TriviaQA questions can be answered through relatively direct retrieval, limiting the additional benefit of fine-grained credit assignment.

\paragraph{Evolution of process signals.}
Figure~\ref{fig:process-ratio} shows the proportions of search turns with positive and negative raw answer-likelihood gains during training.
As training proceeds, the proportion of positively attributed search turns under LOTAPO gradually increases, while the proportion of negatively attributed turns continues to decrease.
IGPO exhibits a similar trend, but its positive proportion grows more slowly than LOTAPO's and it retains more negative process updates late in training.
This change is consistent with the policy gradually increasing positively attributed turns and reducing negatively attributed turns, although turn proportions alone do not establish a causal improvement in specific search behavior.

\subsection{Ablation Study}
\label{sec:ablation}

\paragraph{Effect of backward leave-one-turn attribution.}
Table~\ref{tab:ablation-results} examines the effects of attribution direction and sign-consistency gating.
Without gating, replacing IGPO's forward attribution with LOTAPO's backward leave-one-turn attribution improves average EM from \(0.273\) to \(0.285\).
The gains appear primarily on multi-hop datasets such as HotpotQA, 2Wiki, and MuSiQue.
This pattern is consistent with LOTAPO's central motivation: it evaluates the attribution of a target turn while preserving the downstream context by replacing the entire turn with \texttt{[DELETE]} in the complete search trajectory and comparing the resulting answer likelihoods.

Changing only the attribution direction, however, yields a relatively modest average improvement.
One reason is that standardization can reverse the direction of an attribution signal.
For example, a search turn with a negative raw answer-likelihood gain may still receive a positive normalized advantage if its score is above the group mean.
Directly optimizing with this advantage would make the policy-update direction contradict the raw attribution obtained from the leave-one-turn comparison.

\paragraph{Effect of sign-consistency gating.}
Adding sign-consistency gating to IGPO improves its average EM from \(0.273\) to \(0.282\), indicating that enforcing directional agreement between the normalized advantage and raw attribution can also improve forward attribution.
The effect is clearer for LOTAPO, where gating improves average EM from \(0.285\) to \(0.326\).

This result suggests that the performance gain of full LOTAPO arises neither solely from backward attribution nor simply from filtering process scores.
Backward attribution provides an estimate of turn-level attribution over the complete trajectory, while sign-consistency gating keeps the normalized relative advantage aligned with the direction of the raw attribution.
Together, they allow LOTAPO to use attribution signals that preserve downstream context while maintaining directionally consistent policy-update signals.

The benefit of gating is not fully uniform across datasets.
For example, LOTAPO without gating performs slightly better on HotpotQA than the full method.
Overall, however, full LOTAPO achieves the highest average EM and the best performance on six of the seven datasets.

\begin{table}[H]
\centering
\small
\setlength{\tabcolsep}{4.2pt}
\caption{
\textbf{Ablation results for attribution direction and sign-consistency gating.}
We compare IGPO's forward attribution with LOTAPO's backward leave-one-turn attribution, with and without gating, using exact match (EM).
All variants use Qwen2.5-3B-Instruct~\cite{qwen2024qwen25} with the same training data and optimization configuration.
The best result in each column is shown in bold.
}
\label{tab:ablation-results}
\begin{tabular}{lcccccccc}
\toprule
\multirow{2}{*}{Method}
& \multicolumn{4}{c}{In-Domain}
& \multicolumn{3}{c}{Out-of-Domain}
& \multirow{2}{*}{Avg.} \\
\cmidrule(lr){2-5}
\cmidrule(lr){6-8}
& NQ & TQ & HotpotQA & 2Wiki
& MuSiQue & Bamboogle & PopQA & \\
\midrule
IGPO w/o Gating
& 0.358 & 0.472 & 0.308 & 0.244 & 0.136 & 0.252 & 0.144 & 0.273 \\
IGPO w/ Gating
& 0.348 & 0.466 & 0.296 & 0.264 & 0.160 & 0.260 & 0.182 & 0.282 \\
LOTAPO w/o Gating
& 0.362 & 0.470 & \textbf{0.342} & 0.278 & 0.176 & 0.200 & 0.170 & 0.285 \\
LOTAPO w/ Gating
& \textbf{0.422} & \textbf{0.506} & 0.338 & \textbf{0.320}
& \textbf{0.188} & \textbf{0.304} & \textbf{0.206} & \textbf{0.326} \\
\bottomrule
\end{tabular}
\end{table}

\section{Conclusion}
\label{sec:conclusion}

We presented LOTAPO, a self-generated process-supervision method for multi-turn search reasoning.
Through backward leave-one-turn attribution, LOTAPO compares the gold-answer likelihood under the full context with that under a context in which a single search turn is replaced in its entirety with \texttt{[DELETE]}, thereby estimating the attribution of each search turn under the current context and policy.
It further uses sign-consistency gating to filter directionally unreliable process signals.
LOTAPO requires no additional process reward model, teacher model, verifier, or LLM-as-a-Judge and can be integrated directly into an existing GRPO training framework.
Across seven knowledge-intensive question-answering datasets, LOTAPO outperforms the compared outcome-reward- and step-reward-based baselines, supporting the utility of retrospective turn-level attribution for credit assignment in multi-turn search.

LOTAPO also has several limitations.
Its leave-one-turn score is a context- and policy-dependent counterfactual attribution rather than a formally identified causal effect, and it does not explicitly model interactions among multiple replaced turns.
The current signal also depends on gold answers, and our experiments are limited to the evaluated model scale, local retrieval environment, and relatively short search trajectories.
Future work may investigate more efficient counterfactual approximations, model higher-order interactions among search turns, extend LOTAPO to longer trajectories, web search, and multimodal agents, and reduce its dependence on explicit gold-answer supervision.

\bibliographystyle{lapo_plainnat}
\bibliography{references}

@inproceedings{jin2025searchr1,
  title     = {{Search-R1}: Training {LLMs} to Reason and Leverage Search Engines with Reinforcement Learning},
  author    = {Jin, Bowen and Zeng, Hansi and Yue, Zhenrui and Yoon, Jinsung and Arik, Sercan O. and Wang, Dong and Zamani, Hamed and Han, Jiawei},
  booktitle = {Proceedings of the Second Conference on Language Modeling},
  year      = {2025},
  url       = {https://openreview.net/forum?id=Rwhi91ideu}
}

@article{song2025r1searcher,
  title   = {{R1-Searcher}: Incentivizing the Search Capability in {LLMs} via Reinforcement Learning},
  author  = {Song, Huatong and Jiang, Jinhao and Min, Yingqian and Chen, Jie and Chen, Zhipeng and Zhao, Wayne Xin and Fang, Lei and Wen, Ji-Rong},
  journal = {arXiv preprint arXiv:2503.05592},
  year    = {2025},
  url     = {https://arxiv.org/abs/2503.05592}
}

@article{chen2025research,
  title   = {{ReSearch}: Learning to Reason with Search for {LLMs} via Reinforcement Learning},
  author  = {Chen, Mingyang and Li, Tianpeng and Sun, Haoze and Zhou, Yijie and Zhu, Chenzheng and Yang, Fan and Zhou, Zenan and Chen, Weipeng and Wang, Haofen and Pan, Jeff Z. and Zhang, Wen and Chen, Huajun},
  journal = {arXiv preprint arXiv:2503.19470},
  year    = {2025},
  url     = {https://arxiv.org/abs/2503.19470}
}

@inproceedings{wei2025webagentr1,
  title     = {{WebAgent-R1}: Training Web Agents via End-to-End Multi-Turn Reinforcement Learning},
  author    = {Wei, Zhepei and Yao, Wenlin and Liu, Yao and Zhang, Weizhi and Lu, Qin and Qiu, Liang and Yu, Changlong and Xu, Puyang and Zhang, Chao and Yin, Bing and Yun, Hyokun and Li, Lihong},
  booktitle = {Proceedings of the 2025 Conference on Empirical Methods in Natural Language Processing},
  pages     = {7909--7928},
  year      = {2025},
  publisher = {Association for Computational Linguistics},
  doi       = {10.18653/v1/2025.emnlp-main.401},
  url       = {https://aclanthology.org/2025.emnlp-main.401/}
}

@inproceedings{wu2026mmsearchr1,
  title     = {{MMSearch-R1}: Incentivizing {LMMs} to Search},
  author    = {Wu, Jinming and Deng, Zihao and Li, Wei and Liu, Yiding and You, Bo and Li, Bo and Ma, Zejun and Liu, Ziwei},
  booktitle = {Proceedings of the 64th Annual Meeting of the Association for Computational Linguistics (Volume 1: Long Papers)},
  pages     = {2456--2487},
  year      = {2026},
  publisher = {Association for Computational Linguistics},
  doi       = {10.18653/v1/2026.acl-long.114},
  url       = {https://aclanthology.org/2026.acl-long.114/}
}

@article{tan2026ragr1,
  title     = {{RAG-R1}: Incentivizing the Search and Reasoning Capabilities of {LLMs} Through Multi-Query Parallelism},
  author    = {Tan, Zhiwen and Huang, Jiaming and Wu, Qintong and Zhang, Hongxuan and Zhuang, Chenyi and Gu, Jinjie},
  journal   = {Proceedings of the AAAI Conference on Artificial Intelligence},
  volume    = {40},
  number    = {39},
  pages     = {33187--33195},
  year      = {2026},
  doi       = {10.1609/aaai.v40i39.40603},
  url       = {https://ojs.aaai.org/index.php/AAAI/article/view/40603}
}

@article{lightman2023verify,
  title   = {Let's Verify Step by Step},
  author  = {Lightman, Hunter and Kosaraju, Vineet and Burda, Yura and Edwards, Harri and Baker, Bowen and Lee, Teddy and Leike, Jan and Schulman, John and Sutskever, Ilya and Cobbe, Karl},
  journal = {arXiv preprint arXiv:2305.20050},
  year    = {2023},
  url     = {https://arxiv.org/abs/2305.20050}
}

@inproceedings{wang2024mathshepherd,
  title     = {{Math-Shepherd}: Verify and Reinforce {LLMs} Step-by-step without Human Annotations},
  author    = {Wang, Peiyi and Li, Lei and Shao, Zhihong and Xu, Runxin and Dai, Damai and Li, Yifei and Chen, Deli and Wu, Yu and Sui, Zhifang},
  booktitle = {Proceedings of the 62nd Annual Meeting of the Association for Computational Linguistics (Volume 1: Long Papers)},
  pages     = {9426--9439},
  year      = {2024},
  publisher = {Association for Computational Linguistics},
  doi       = {10.18653/v1/2024.acl-long.510},
  url       = {https://aclanthology.org/2024.acl-long.510/}
}

@article{uesato2022processfeedback,
  title   = {Solving Math Word Problems with Process- and Outcome-Based Feedback},
  author  = {Uesato, Jonathan and Kushman, Nate and Kumar, Ramana and Song, Francis and Siegel, Noah and Wang, Lisa and Creswell, Antonia and Irving, Geoffrey and Higgins, Irina},
  journal = {arXiv preprint arXiv:2211.14275},
  year    = {2022},
  url     = {https://arxiv.org/abs/2211.14275}
}

@inproceedings{zhang2024restmcts,
  title     = {{ReST-MCTS*}: {LLM} Self-Training via Process Reward Guided Tree Search},
  author    = {Zhang, Dan and Zhoubian, Sining and Hu, Ziniu and Yue, Yisong and Dong, Yuxiao and Tang, Jie},
  booktitle = {Advances in Neural Information Processing Systems},
  volume    = {37},
  pages     = {64735--64772},
  year      = {2024},
  doi       = {10.52202/079017-2066},
  url       = {https://proceedings.neurips.cc/paper_files/paper/2024/hash/76ec4dc30e9faaf0e4b6093eaa377218-Abstract-Conference.html}
}

@article{fang2026ropd,
  title   = {Rubric-based On-policy Distillation},
  author  = {Fang, Junfeng and Hong, Zhepei and Zheng, Mao and Song, Mingyang and Li, Gengsheng and Jiang, Houcheng and Zhang, Dan and Guo, Haiyun and Wang, Xiang and Chua, Tat-Seng},
  journal = {arXiv preprint arXiv:2605.07396},
  year    = {2026},
  url     = {https://arxiv.org/abs/2605.07396}
}

@article{yang2026opid,
  title   = {{OPID}: On-Policy Skill Distillation for Agentic Reinforcement Learning},
  author  = {Yang, Shuo and Wu, Jinyang and Lu, Zhengxi and Shen, Yuhao and Zhang, Fan and Feng, Lang and Zhang, Shuai and Luo, Haoran and Lian, Zheng and Wen, Zhengqi and Tao, Jianhua},
  journal = {arXiv preprint arXiv:2606.26790},
  year    = {2026},
  url     = {https://arxiv.org/abs/2606.26790}
}

@inproceedings{wang2026igpo,
  title     = {Information Gain-based Policy Optimization: A Simple and Effective Approach for Multi-Turn Search Agents},
  author    = {Wang, Guoqing and Dai, Sunhao and Ye, Guangze and Gan, Zeyu and Yao, Wei and Deng, Yong and Wu, Xiaofeng and Ying, Zhenzhe},
  booktitle = {The Fourteenth International Conference on Learning Representations},
  year      = {2026},
  url       = {https://openreview.net/forum?id=qkWP6phrvZ}
}

@inproceedings{xie2026tips,
  title     = {{TIPS}: Turn-Level Information-Potential Reward Shaping for Search-Augmented {LLMs}},
  author    = {Xie, Yutao and Thomas, Nathaniel and Hansen, Nicklas and Fu, Yang and Li, Li Erran and Wang, Xiaolong},
  booktitle = {The Fourteenth International Conference on Learning Representations},
  year      = {2026},
  url       = {https://openreview.net/forum?id=eBMOr6a84z}
}

@inproceedings{zheng2025stepsearch,
  title     = {{StepSearch}: Igniting {LLMs} Search Ability via Step-Wise Proximal Policy Optimization},
  author    = {Zheng, Xuhui and An, Kang and Wang, Ziliang and Wang, Yuhang and Wu, Yichao},
  booktitle = {Proceedings of the 2025 Conference on Empirical Methods in Natural Language Processing},
  pages     = {21805--21830},
  year      = {2025},
  publisher = {Association for Computational Linguistics},
  doi       = {10.18653/v1/2025.emnlp-main.1106},
  url       = {https://aclanthology.org/2025.emnlp-main.1106/}
}

@article{liang2026igsearch,
  title   = {{IG-Search}: Step-Level Information Gain Rewards for Search-Augmented Reasoning},
  author  = {Liang, Zihan and Ma, Yufei and Chen, Ben and Qian, Zhipeng and Dai, Huangyu and Mao, Lingtao and Zhang, Xuxin and Lei, Chenyi and Ou, Wenwu},
  journal = {arXiv preprint arXiv:2604.15148},
  year    = {2026},
  url     = {https://arxiv.org/abs/2604.15148}
}

@article{shao2024deepseekmath,
  title   = {{DeepSeekMath}: Pushing the Limits of Mathematical Reasoning in Open Language Models},
  author  = {Shao, Zhihong and Wang, Peiyi and Zhu, Qihao and Xu, Runxin and Song, Junxiao and Bi, Xiao and Zhang, Haowei and Zhang, Mingchuan and Li, Y. K. and Wu, Y. and Guo, Daya},
  journal = {arXiv preprint arXiv:2402.03300},
  year    = {2024},
  url     = {https://arxiv.org/abs/2402.03300}
}

@inproceedings{li2025searcho1,
  title     = {{Search-o1}: Agentic Search-Enhanced Large Reasoning Models},
  author    = {Li, Xiaoxi and Dong, Guanting and Jin, Jiajie and Zhang, Yuyao and Zhou, Yujia and Zhu, Yutao and Zhang, Peitian and Dou, Zhicheng},
  booktitle = {Proceedings of the 2025 Conference on Empirical Methods in Natural Language Processing},
  pages     = {5420--5438},
  year      = {2025},
  publisher = {Association for Computational Linguistics},
  doi       = {10.18653/v1/2025.emnlp-main.276},
  url       = {https://aclanthology.org/2025.emnlp-main.276/}
}

@inproceedings{wei2022chainofthought,
  title     = {Chain-of-Thought Prompting Elicits Reasoning in Large Language Models},
  author    = {Wei, Jason and Wang, Xuezhi and Schuurmans, Dale and Bosma, Maarten and Ichter, Brian and Xia, Fei and Chi, Ed and Le, Quoc V. and Zhou, Denny},
  booktitle = {Advances in Neural Information Processing Systems},
  volume    = {35},
  pages     = {24824--24837},
  year      = {2022},
  url       = {https://proceedings.neurips.cc/paper/2022/hash/9d5609613524ecf4f15af0f7b31abca4-Abstract-Conference.html}
}

@inproceedings{lewis2020rag,
  title     = {Retrieval-Augmented Generation for Knowledge-Intensive {NLP} Tasks},
  author    = {Lewis, Patrick and Perez, Ethan and Piktus, Aleksandra and Petroni, Fabio and Karpukhin, Vladimir and Goyal, Naman and K{\"u}ttler, Heinrich and Lewis, Mike and Yih, Wen-tau and Rockt{\"a}schel, Tim and Riedel, Sebastian and Kiela, Douwe},
  booktitle = {Advances in Neural Information Processing Systems},
  volume    = {33},
  pages     = {9459--9474},
  year      = {2020},
  url       = {https://proceedings.neurips.cc/paper/2020/hash/6b493230205f780e1bc26945df7481e5-Abstract.html}
}

@article{kwiatkowski2019naturalquestions,
  title   = {Natural Questions: A Benchmark for Question Answering Research},
  author  = {Kwiatkowski, Tom and Palomaki, Jennimaria and Redfield, Olivia and Collins, Michael and Parikh, Ankur and Alberti, Chris and Epstein, Danielle and Polosukhin, Illia and Devlin, Jacob and Lee, Kenton and Toutanova, Kristina and Jones, Llion and Kelcey, Matthew and Chang, Ming-Wei and Dai, Andrew M. and Uszkoreit, Jakob and Le, Quoc and Petrov, Slav},
  journal = {Transactions of the Association for Computational Linguistics},
  volume  = {7},
  pages   = {452--466},
  year    = {2019},
  doi     = {10.1162/tacl_a_00276},
  url     = {https://aclanthology.org/Q19-1026/}
}

@inproceedings{joshi2017triviaqa,
  title     = {{TriviaQA}: A Large Scale Distantly Supervised Challenge Dataset for Reading Comprehension},
  author    = {Joshi, Mandar and Choi, Eunsol and Weld, Daniel S. and Zettlemoyer, Luke},
  booktitle = {Proceedings of the 55th Annual Meeting of the Association for Computational Linguistics (Volume 1: Long Papers)},
  pages     = {1601--1611},
  year      = {2017},
  publisher = {Association for Computational Linguistics},
  doi       = {10.18653/v1/P17-1147},
  url       = {https://aclanthology.org/P17-1147/}
}

@inproceedings{yang2018hotpotqa,
  title     = {{HotpotQA}: A Dataset for Diverse, Explainable Multi-hop Question Answering},
  author    = {Yang, Zhilin and Qi, Peng and Zhang, Saizheng and Bengio, Yoshua and Cohen, William W. and Salakhutdinov, Ruslan and Manning, Christopher D.},
  booktitle = {Proceedings of the 2018 Conference on Empirical Methods in Natural Language Processing},
  pages     = {2369--2380},
  year      = {2018},
  publisher = {Association for Computational Linguistics},
  doi       = {10.18653/v1/D18-1259},
  url       = {https://aclanthology.org/D18-1259/}
}

@inproceedings{ho2020twowiki,
  title     = {Constructing A Multi-hop {QA} Dataset for Comprehensive Evaluation of Reasoning Steps},
  author    = {Ho, Xanh and Duong Nguyen, Anh-Khoa and Sugawara, Saku and Aizawa, Akiko},
  booktitle = {Proceedings of the 28th International Conference on Computational Linguistics},
  pages     = {6609--6625},
  year      = {2020},
  publisher = {International Committee on Computational Linguistics},
  doi       = {10.18653/v1/2020.coling-main.580},
  url       = {https://aclanthology.org/2020.coling-main.580/}
}

@article{trivedi2022musique,
  title   = {{MuSiQue}: Multihop Questions via Single-hop Question Composition},
  author  = {Trivedi, Harsh and Balasubramanian, Niranjan and Khot, Tushar and Sabharwal, Ashish},
  journal = {Transactions of the Association for Computational Linguistics},
  volume  = {10},
  pages   = {539--554},
  year    = {2022},
  doi     = {10.1162/tacl_a_00475},
  url     = {https://aclanthology.org/2022.tacl-1.31/}
}

@inproceedings{press2023compositionality,
  title     = {Measuring and Narrowing the Compositionality Gap in Language Models},
  author    = {Press, Ofir and Zhang, Muru and Min, Sewon and Schmidt, Ludwig and Smith, Noah A. and Lewis, Mike},
  booktitle = {Findings of the Association for Computational Linguistics: EMNLP 2023},
  pages     = {5687--5711},
  year      = {2023},
  publisher = {Association for Computational Linguistics},
  doi       = {10.18653/v1/2023.findings-emnlp.378},
  url       = {https://aclanthology.org/2023.findings-emnlp.378/}
}

@inproceedings{mallen2023popqa,
  title     = {When Not to Trust Language Models: Investigating Effectiveness of Parametric and Non-Parametric Memories},
  author    = {Mallen, Alex and Asai, Akari and Zhong, Victor and Das, Rajarshi and Khashabi, Daniel and Hajishirzi, Hannaneh},
  booktitle = {Proceedings of the 61st Annual Meeting of the Association for Computational Linguistics (Volume 1: Long Papers)},
  pages     = {9802--9822},
  year      = {2023},
  publisher = {Association for Computational Linguistics},
  doi       = {10.18653/v1/2023.acl-long.546},
  url       = {https://aclanthology.org/2023.acl-long.546/}
}

@inproceedings{karpukhin2020dense,
  title     = {Dense Passage Retrieval for Open-Domain Question Answering},
  author    = {Karpukhin, Vladimir and Oguz, Barlas and Min, Sewon and Lewis, Patrick and Wu, Ledell and Edunov, Sergey and Chen, Danqi and Yih, Wen-tau},
  booktitle = {Proceedings of the 2020 Conference on Empirical Methods in Natural Language Processing},
  pages     = {6769--6781},
  year      = {2020},
  publisher = {Association for Computational Linguistics},
  doi       = {10.18653/v1/2020.emnlp-main.550},
  url       = {https://aclanthology.org/2020.emnlp-main.550/}
}

@article{wang2022e5,
  title   = {Text Embeddings by Weakly-Supervised Contrastive Pre-training},
  author  = {Wang, Liang and Yang, Nan and Huang, Xiaolong and Jiao, Binxing and Yang, Linjun and Jiang, Daxin and Majumder, Rangan and Wei, Furu},
  journal = {arXiv preprint arXiv:2212.03533},
  year    = {2022},
  url     = {https://arxiv.org/abs/2212.03533}
}

@article{qwen2024qwen25,
  title   = {{Qwen2.5} Technical Report},
  author  = {{Qwen} and Yang, An and Yang, Baosong and Zhang, Beichen and Hui, Binyuan and Zheng, Bo and Yu, Bowen and Li, Chengyuan and Liu, Dayiheng and Huang, Fei and Wei, Haoran and Lin, Huan and Yang, Jian and Tu, Jianhong and Zhang, Jianwei and Yang, Jianxin and Yang, Jiaxi and Zhou, Jingren and Lin, Junyang and Dang, Kai and Lu, Keming and Bao, Keqin and Yang, Kexin and Yu, Le and Li, Mei and Xue, Mingfeng and Zhang, Pei and Zhu, Qin and Men, Rui and Lin, Runji and Li, Tianhao and Tang, Tianyi and Xia, Tingyu and Ren, Xingzhang and Ren, Xuancheng and Fan, Yang and Su, Yang and Zhang, Yichang and Wan, Yu and Liu, Yuqiong and Cui, Zeyu and Zhang, Zhenru and Qiu, Zihan},
  journal = {arXiv preprint arXiv:2412.15115},
  year    = {2024},
  url     = {https://arxiv.org/abs/2412.15115}
}

@article{hu2026inforeasoner,
  title   = {Optimizing Agentic Reasoning with Retrieval via Synthetic Semantic Information Gain Reward},
  author  = {Hu, Senkang and Dai, Yong and Zhao, Yuzhi and Tao, Yihang and Guo, Yu and Fang, Zhengru and Kwong, Sam Tak Wu and Fang, Yuguang},
  journal = {arXiv preprint arXiv:2602.00845},
  year    = {2026},
  note    = {Accepted at the 43rd International Conference on Machine Learning},
  url     = {https://arxiv.org/abs/2602.00845}
}

@inproceedings{zhang2026criticsearch,
  title     = {{CriticSearch}: Fine-Grained Credit Assignment for Search Agents via a Retrospective Critic},
  author    = {Zhang, Yaocheng and Huang, Haohuan and Song, Zijun and Zhu, Yuanheng and Zhang, Qichao and Zhao, Zijie and Zhao, Dongbin},
  booktitle = {Findings of the Association for Computational Linguistics: ACL 2026},
  pages     = {12272--12290},
  year      = {2026},
  publisher = {Association for Computational Linguistics},
  doi       = {10.18653/v1/2026.findings-acl.596},
  url       = {https://aclanthology.org/2026.findings-acl.596/}
}

@article{feng2026rewardflow,
  title   = {{RewardFlow}: Topology-Aware Reward Propagation on State Graphs for Agentic {RL} with Large Language Models},
  author  = {Feng, Xiao and Han, Bo and Zhou, Zhanke and Fan, Jiaqi and Yao, Jiangchao and Li, Ka Ho and Yu, Dahai and Ng, Michael Kwok-Po},
  journal = {arXiv preprint arXiv:2603.18859},
  year    = {2026},
  url     = {https://arxiv.org/abs/2603.18859}
}

@article{wei2025turnlevel,
  title   = {Reinforcing Multi-Turn Reasoning in {LLM} Agents via Turn-Level Reward Design},
  author  = {Wei, Quan and Zeng, Siliang and Li, Chenliang and Brown, William and Frunza, Oana and Deng, Wei and Schneider, Anderson and Nevmyvaka, Yuriy and Zhao, Yang Katie and Garcia, Alfredo and Hong, Mingyi},
  journal = {arXiv preprint arXiv:2505.11821},
  year    = {2025},
  url     = {https://arxiv.org/abs/2505.11821}
}

\appendix

\section{Reproducibility Details}
\label{app:implementation_details}

This section provides additional details about LOTAPO's training configuration, retrieval environment, and counterfactual scoring implementation.
Unless otherwise specified, all reinforcement-learning methods use the same model initialization, training data, prompt template, decoding parameters, and retrieval budget.

\subsection{Training and Optimization Configuration}

Table~\ref{tab:appendix-training-config} summarizes the key hyperparameters for training and LOTAPO process scoring.
The base policy is initialized from Qwen2.5-3B-Instruct and optimized with GRPO's group-relative advantage estimation and clipped surrogate objective.
We use the AdamW optimizer, and all experiments are conducted on a single node with eight NVIDIA A100 GPUs.

\begin{table}[t]
\centering
\small
\setlength{\tabcolsep}{5pt}
\caption{Key configurations for training and LOTAPO process scoring.}
\label{tab:appendix-training-config}
\begin{tabular}{ll}
\toprule
Configuration & Value \\
\midrule
Global training batch size & $512$ \\
PPO mini-batch size & $256$ \\
PPO micro-batch size & $64$ \\
Sampled trajectories per question $G$ & $5$ \\
Optimizer & AdamW \\
Learning rate & $1\times10^{-6}$ \\
PPO clipping range $\epsilon_{\mathrm{clip}}$ & $0.2$ \\
KL coefficient & $0.001$ \\
Optimization steps & $200$ \\
Training hardware & $8\times$ NVIDIA A100 \\
Process-advantage weight $\lambda$ & $0.5$ \\
Answer-likelihood-gain stability term $\epsilon_{\mathrm{ig}}$ (\texttt{ig\_eps}) & $10^{-6}$ \\
\bottomrule
\end{tabular}
\end{table}

\subsection{Retrieval Environment and Generation Settings}

All methods share the same local retrieval environment.
The retrieval corpus is constructed from the 2018 English Wikipedia, and similarity search is performed using an E5 dense retriever with a FAISS index.
Each search returns the three most relevant passages.
The retrieval results are wrapped in a unified title-and-body format inside \texttt{<information>} tags and appended to the current context as environment-generated observations.

The model may perform at most three search interactions.
After exhausting the search budget or generating a final response, the trajectory enters or completes the answer-generation stage.
The final-answer turn does not trigger retrieval and does not participate in LOTAPO process attribution.
We use stochastic sampling during training and greedy decoding for validation and testing.
Table~\ref{tab:appendix-generation-config} provides the detailed configuration.

\begin{table}[t]
\centering
\small
\setlength{\tabcolsep}{5pt}
\caption{Key retrieval and generation configurations.}
\label{tab:appendix-generation-config}
\begin{tabular}{ll}
\toprule
Configuration & Value \\
\midrule
Passages returned per retrieval & $3$ \\
Maximum search turns & $3$ \\
Context length & $4096$ tokens \\
Maximum initial input length & $2048$ tokens \\
Maximum tokens per generation & $500$ tokens \\
Maximum length per retrieval observation & $500$ tokens \\
Training decoding temperature & $1.0$ \\
Validation and test decoding & Greedy decoding \\
\bottomrule
\end{tabular}
\end{table}

\subsection{Counterfactual-Context Construction and Batched Scoring}

For each valid search turn $t$ in trajectory $i$, we replace the entire search action and its retrieval observation with a fixed \texttt{[DELETE]} marker to obtain the counterfactual context $c_{i,-t}$.
This operation preserves the order of all interactions before and after the target turn and explicitly marks the absence of the target interaction with a placeholder.
The design avoids directly concatenating the text before and after the target turn, although we do not separately evaluate the placeholder's effect on contextual coherence.
The final-answer turn is always excluded from the scoring context, so the gold-answer likelihood reflects only the support provided by preceding search interactions.

To compute answer-likelihood gains, we flatten the full contexts and all leave-one-turn counterfactual contexts in the same rollout batch into a single scoring batch.
We append the same gold-answer scoring template to every context and compute the mean log-likelihood only over the token span corresponding to the gold answer.
The full-context score is reused across all turns in the same trajectory, reducing redundant forward computation and model-scheduling overhead.

\subsection{Training Cost}

\begin{figure}[t]
\centering
\includegraphics[width=0.82\linewidth]{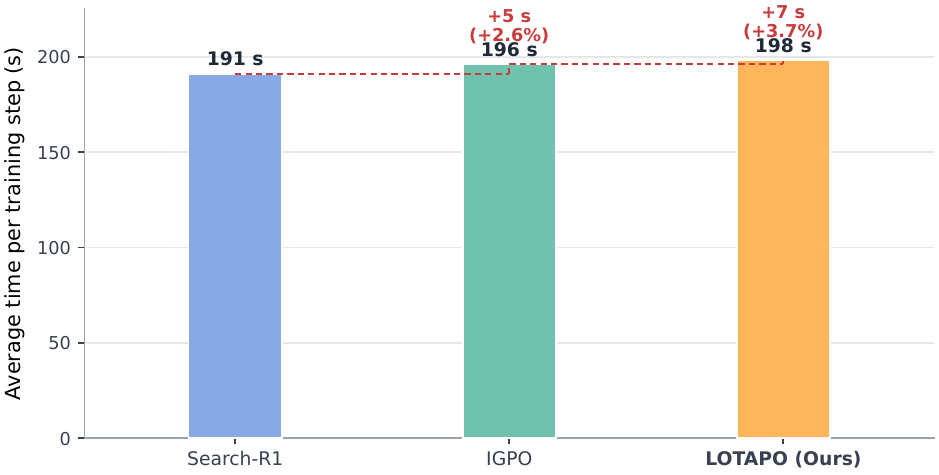}
\caption{\textbf{Training time per step for different methods.}
Under the same hardware environment and training configuration, the average training times per step for Search-R1, IGPO, and LOTAPO are 191, 196, and 198 seconds, respectively.
The red dashed lines and annotations indicate the additional wall-clock time and relative increase of IGPO and LOTAPO over Search-R1.}
\label{fig:appendix-compute-budget}
\end{figure}

To assess the additional training cost of counterfactual process attribution, we measure the average training time per step for different methods under the same hardware environment and training configuration.
The results are shown in Figure~\ref{fig:appendix-compute-budget}.

Search-R1 requires an average of 191 seconds per training step.
Introducing turn-level process signals increases this time to 196 seconds for IGPO, an overhead of 5 seconds (\(2.6\%\)).
LOTAPO requires 198 seconds per step, only 7 seconds (\(3.7\%\)) more than Search-R1 and 2 seconds (approximately \(1.0\%\)) more than IGPO.
Under the current implementation and experimental configuration, backward leave-one-turn attribution therefore does not produce a large increase in training time.
Its additional overhead is controlled mainly by batching the full and counterfactual contexts for scoring and reusing the full-context score across turns in the same trajectory.
These measurements are wall-clock times for specific hardware and implementation choices, not platform-independent theoretical computational costs.

\FloatBarrier
\section{Prompt Template}
\label{app:prompt_templates}

\subsection{Agent Interaction Format}
\label{sec:agent_interaction_format}

For each input question, we construct the model input with a fixed task prompt and encode it in conversation format using the base model's native chat template, with an assistant generation prompt appended at the end.
The complete task prompt is:

\begin{Verbatim}[fontsize=\small,breaklines=true,breakanywhere=true]
You are answering a question that may require external search.

At each step, choose exactly one action:

1. If external information is needed:
<think>Write your brief reasoning about what information is
missing and why search is needed.</think>
<search>one concise search query</search>

2. If enough information is available:
<think>Write your brief reasoning leading to the answer.</think>
<answer>final answer only</answer>

Rules:
- Do not output the words "reasoning", "query", or
  "final answer" as placeholders.
- Use either <search> or <answer>, never both.
- The <search> tag must contain only one concise query.
- Search results may be provided as
  <information>...</information>; read them but never
  generate <information>.
- The <answer> tag should contain only the final answer,
  with no explanation.
- Stop immediately after </search> or </answer>.

Question: {question}
\end{Verbatim}

At each interaction turn, the policy first produces brief reasoning for the current step inside \texttt{<think>} and then selects either a search action or an answer action.
When the available information is insufficient, the policy outputs:

\begin{Verbatim}[fontsize=\small]
<think>brief reasoning for search</think>
<search>one concise search query</search>
\end{Verbatim}

The \texttt{<search>} tag contains only one query to be sent to the retrieval service.
When the policy determines that the context contains sufficient information, it outputs:

\begin{Verbatim}[fontsize=\small]
<think>brief reasoning leading to the answer</think>
<answer>final answer only</answer>
\end{Verbatim}

The \texttt{<answer>} tag contains only the final answer, without additional explanation.

Each model call may generate exactly one action, either a search action or an answer action, but not both.
Decoding terminates immediately after the first complete and valid \texttt{</search>} or \texttt{</answer>} tag, preventing the model from executing multiple actions in a single generation.

When the policy generates a search action, the environment extracts the query from the \texttt{<search>} tag and invokes the external retrieval service.
The environment wraps the retrieved passages as:

\begin{Verbatim}[fontsize=\small]
<information>retrieved passages</information>
\end{Verbatim}

This retrieval observation is then appended to the current conversation context and used as input to the next model generation.
The \texttt{<information>} tag is generated only by the environment; the policy never generates it.
When the policy generates an answer action, the current trajectory terminates immediately.

A complete multi-turn search trajectory can therefore be represented as:

\begin{Verbatim}[fontsize=\small]
<think>...</think>
<search>...</search>
<information>...</information>
...
<think>...</think>
<search>...</search>
<information>...</information>
<think>...</think>
<answer>...</answer>
\end{Verbatim}

A trajectory contains zero or more
\texttt{<think>-<search>-<information>} interactions and ends with one
\texttt{<think>-<answer>} interaction.

\subsection{Gold-Answer Likelihood Scoring Template}

LOTAPO appends the same fixed response template to both the full and counterfactual contexts:

\begin{Verbatim}[fontsize=\small]
<think>Now there's enough information to answer</think>
<answer>{gold_answer}</answer>
\end{Verbatim}

During scoring, we sum only the log probabilities of the tokens corresponding to \texttt{\{gold\_answer\}} and average them over the number of answer tokens.
The reasoning prefix, answer tags, and closing tag in the template are excluded from the score.
This design compares the full and counterfactual contexts under the same target sequence and reduces the effect of answer-length differences on the attribution scale.

\FloatBarrier
\section{Data Processing and Evaluation Details}
\label{app:data_and_evaluation}

\subsection{Dataset Construction}

The training set consists of the training splits of NQ, TriviaQA, HotpotQA, and 2Wiki, with 20,000 questions randomly sampled from each dataset and 80,000 training examples in total.
Validation and testing cover these four in-domain datasets and the three out-of-domain datasets MuSiQue, Bamboogle, and PopQA, with testing conducted on the corresponding test or validation split of each dataset.
Table~\ref{tab:appendix-data-splits} summarizes the training and validation sample sizes.

\begin{table}[t]
\centering
\small
\setlength{\tabcolsep}{7pt}
\caption{Sample sizes for training and validation. A dash indicates that the dataset is not used for reinforcement-learning training.}
\label{tab:appendix-data-splits}
\begin{tabular}{lcc}
\toprule
Dataset & Training Examples & Validation Examples \\
\midrule
Natural Questions & $20{,}000$ & $500$ \\
TriviaQA & $20{,}000$ & $500$ \\
HotpotQA & $20{,}000$ & $500$ \\
2WikiMultiHopQA & $20{,}000$ & $500$ \\
MuSiQue & -- & $500$ \\
Bamboogle & -- & $500$ \\
PopQA & -- & $500$ \\
\bottomrule
\end{tabular}
\end{table}

Each example contains a question in conversation format and one or more reference answers.
When multiple reference answers are available, we compute both the training reward and test metric against every reference and take the best match.

\subsection{Answer Extraction, Normalization, and Evaluation Metrics}

Before scoring, we extract the last complete \texttt{<answer>} block from the model output as the predicted answer.
We normalize answers by lowercasing, removing ASCII punctuation, removing the English articles \texttt{a/an/the}, and collapsing consecutive whitespace, in that order.

During training, answer-level F1 is used as the terminal task reward.
Specifically, we tokenize the normalized prediction and reference answer on whitespace, compute precision and recall from the multiset intersection of tokens, and take their harmonic mean.
At test time, exact match (EM) is the primary metric: a prediction receives 1 if its normalized form exactly matches any normalized reference answer and 0 otherwise.
The average results reported in the main paper are macro-average EM scores across the seven datasets, with equal weight assigned to every dataset.

\FloatBarrier

\section{Case Studies}
\label{app:case-study}

This section uses two concrete cases to illustrate two key components of LOTAPO.
First, backward leave-one-turn attribution evaluates a target search turn while preserving all downstream interactions, allowing it to identify delayed attribution that becomes apparent only through subsequent reasoning and evidence integration.
Second, sign-consistency gating filters normalized process signals that lack support from the raw counterfactual attribution, preventing the policy from rewarding or penalizing turns with no observed attribution.
Unless otherwise specified, all values are rounded to three decimal places.

\subsection{Backward Attribution Identifies Delayed Cross-Turn Contributions}
\label{app:case-direction}

Figure~\ref{fig:case-delayed-contribution} presents a representative example.
The target search turn retrieves evidence directly related to the gold answer, but introducing this evidence does not immediately increase the likelihood of generating that answer.
Forward attribution, which compares adjacent prefixes before and after the target turn, therefore assigns it negative attribution.
In contrast, LOTAPO evaluates the turn retrospectively within the complete trajectory and identifies the evidence's positive role in subsequent reasoning and evidence integration.

\begin{figure*}[t]
\centering
\small

\begin{minipage}[t]{0.57\textwidth}
\vspace{0pt}
\textbf{Example Trajectory}

\vspace{1mm}
\renewcommand{\arraystretch}{1.18}
\begin{tabular}{
    @{}
    p{0.22\linewidth}
    p{0.72\linewidth}
    @{}
}
\toprule
\textbf{Stage}
&
\textbf{Content}
\\
\midrule

Question
&
Who is the main character in \emph{Green Eggs and Ham}?
\\

Gold answer
&
\texttt{Sam-I-am}
\\

Initial search
&
\texttt{main character in Green Eggs and Ham}
\\

\rowcolor{lapogreen!8}
Target turn
&
The retrieved passage states that \textbf{Sam-I-Am}
repeatedly persuades another character to try green eggs and ham.
\\

Later interaction
&
The model distinguishes Sam-I-Am from the unnamed character
and identifies him as the character driving the main plot.
\\

Final answer
&
\texttt{Sam-I-am}
\\

\bottomrule
\end{tabular}
\end{minipage}
\hfill
\begin{minipage}[t]{0.39\textwidth}
\vspace{0pt}
\textbf{Attribution Comparison}

\vspace{1mm}
\footnotesize
\setlength{\tabcolsep}{2pt}
\renewcommand{\arraystretch}{1.22}
\begin{tabular}{
    @{}
    p{0.22\linewidth}
    p{0.17\linewidth}
    p{0.18\linewidth}
    p{0.31\linewidth}
    @{}
}
\toprule
\textbf{Method}
&
$\boldsymbol{\Delta_t}$
&
$\boldsymbol{\widetilde{z}_t}$
&
\textbf{Result}
\\
\midrule

\rowcolor{lapored!7}
Forward
&
$-1.858$
&
$-2.149$
&
Negative
\\

\rowcolor{lapogreen!8}
LOTAPO
&
$+1.556$
&
$+1.875$
&
Positive
\\

\bottomrule
\end{tabular}

\vspace{3mm}

\noindent
\textbf{Interpretation.}
Forward attribution measures only the immediate change after the target turn is added and therefore assigns it negative attribution.
LOTAPO evaluates the target turn with the complete downstream interaction preserved and identifies its delayed positive attribution.
\end{minipage}

\caption{
A case in which backward leave-one-turn attribution identifies a delayed cross-turn contribution.
The target search turn retrieves key evidence supporting the gold answer \texttt{Sam-I-am}, but the likelihood of generating the gold answer temporarily decreases when the evidence is introduced, leading forward attribution to assign a negative value.
LOTAPO compares the full context with a counterfactual context in which the target turn is replaced in its entirety with \texttt{[DELETE]}.
By preserving the complete search and reasoning context after the target turn, LOTAPO identifies the evidence's positive attribution within the current trajectory.
}
\label{fig:case-delayed-contribution}
\end{figure*}

This case illustrates that an immediate change in answer confidence caused by a search turn is not equivalent to the turn's overall attribution to the final reasoning outcome.
New evidence may temporarily perturb the model's answer distribution but later play an important role in further search, entity disambiguation, and evidence integration.
Forward attribution based only on adjacent prefixes cannot observe this delayed effect and may therefore assign negative attribution to a search turn with longer-term value.

LOTAPO preserves all interactions after the target turn and constructs a counterfactual context by replacing that turn in its entirety with \texttt{[DELETE]}.
If the replacement lowers the gold-answer likelihood, the target turn receives positive attribution under the current full context and policy.
Backward attribution can therefore evaluate cross-turn dependencies retrospectively and reduce errors caused by local-prefix comparisons when attribution is delayed.

\subsection{Sign-Consistency Gating Filters Directional Distortions from Normalization}
\label{app:case-gating}

Group normalization constructs a process signal from the contribution of each search turn relative to the group mean.
Consequently, even when a turn has zero raw counterfactual attribution, its normalized score may be positive or negative.
Using this normalized score directly for policy optimization could reward a search turn with no observed attribution or penalize another such turn.
Table~\ref{tab:case-gating} illustrates both cases.

\begin{table}[t]
\centering
\small
\renewcommand{\arraystretch}{1.16}
\setlength{\tabcolsep}{4.5pt}

\caption{
Sign-consistency gating filters normalized process signals that lack support from the raw counterfactual attribution.
$A_t^{\mathrm{proc}}$ denotes the gated process advantage,
$A_t^{\mathrm{no\ gate}}=\widetilde{z}_t$ denotes the process advantage without gating, and
$\lambda A_t^{\mathrm{no\ gate}}$ and $\lambda A_t^{\mathrm{proc}}$
denote the auxiliary advantages actually added to the policy-generated tokens of the target search turn without and with gating, respectively.
}
\label{tab:case-gating}

\begin{tabular}{
    @{}
    p{0.27\linewidth}
    c
    c
    c
    c
    c
    @{}
}
\toprule
\textbf{Case}
&
$\boldsymbol{\Delta_t}$
&
$\boldsymbol{\widetilde{z}_t}$
&
$\boldsymbol{A_t^{\mathrm{proc}}}$
&
$\boldsymbol{\lambda A_t^{\mathrm{no\ gate}}}$
&
$\boldsymbol{\lambda A_t^{\mathrm{proc}}}$
\\
\midrule

\rowcolor{lapoorange!7}
\emph{Waiting for a Girl Like You}
&
$0.000$
&
$+0.550$
&
$0.000$
&
$+0.275$
&
$0.000$
\\

\rowcolor{lapored!6}
Charlie Bucket actor
&
$0.000$
&
$-0.558$
&
$0.000$
&
$-0.279$
&
$0.000$
\\

\bottomrule
\end{tabular}
\end{table}

In the first case, the target turn has a raw answer-likelihood gain of zero, but group normalization produces a positive score.
Without gating, the process advantage for this turn would be \(+0.550\).
The second case exhibits the opposite pattern: the raw attribution is again zero, but normalization produces a negative score, yielding a process advantage of \(-0.558\) without gating.
Neither update direction is supported by the raw counterfactual attribution.
In the table, the gated process advantages are zero in both cases; after multiplication by the process-advantage weight \(\lambda=0.5\), the auxiliary advantages added to the policy-generated tokens also remain zero.
Without gating, the corresponding auxiliary advantages would instead be \(0.550\times0.5=0.275\) and \(-0.558\times0.5=-0.279\).

To filter such directionally distorted process signals, LOTAPO defines sign-consistency gating as

\begin{equation}
\label{eq:case-sign-gating}
g_{i,t}
=
\mathbb{I}
\left[
\Delta_{i,t}\widetilde{z}_{i,t}>0
\right],
\end{equation}

where $\Delta_{i,t}$ is the raw answer-likelihood gain for search turn $t$ in trajectory $i$, and
$\widetilde{z}_{i,t}$ is the process signal after robust scaling, the tanh bounded transformation, and group normalization, in that order.
The gated process advantage is

\begin{equation}
\label{eq:case-gated-advantage}
A_{i,t}^{\mathrm{proc}}
=
g_{i,t}\widetilde{z}_{i,t}.
\end{equation}

The auxiliary advantage actually added to the policy-generated tokens of the search turn is
\(\lambda A_{i,t}^{\mathrm{proc}}\), consistent with Eq.~\eqref{eq:token_advantage}.

Thus, the auxiliary process advantage is retained only when the raw counterfactual attribution and normalized process signal have the same nonzero direction.
When the raw attribution is zero or normalization reverses the sign of the signal, the gate is zero and the corresponding turn receives no auxiliary process update.

Importantly, sign-consistency gating filters only fine-grained process advantages that lack counterfactual support.
It does not remove the trajectory-level outcome advantage or alter the outcome supervision received by the final-answer turn.

\end{document}